\documentclass[10pt,twocolumn,letterpaper]{article}

\usepackage{cvpr}              % To produce the CAMERA-READY version
\usepackage{graphicx}
\usepackage{amsmath}
\usepackage{amssymb}
\usepackage{booktabs}
\usepackage{multirow}
\usepackage{pifont}
\usepackage{colortbl}
\definecolor{Gray}{gray}{0.9}
\newcommand{\cmark}{\ding{51}}%
\usepackage[ruled]{algorithm2e}
\usepackage{bm}

\usepackage[pagebackref,breaklinks,colorlinks,bookmarks=false]{hyperref}

\usepackage[capitalize]{cleveref}
\crefname{section}{Sec.}{Secs.}
\Crefname{section}{Section}{Sections}
\Crefname{table}{Table}{Tables}
\crefname{table}{Tab.}{Tabs.}

\def\cvprPaperID{7761} % *** Enter the CVPR Paper ID here
\def\confName{CVPR}
\def\confYear{2022}

\begin{document}

% \title{Building Robust Multimodal Transformer to Missing Modality}
% \title{Building Multimodal Transformer with Robustness to Missing Modality}
\title{Are Multimodal Transformers Robust to Missing Modality?}
\author{
Mengmeng Ma\textsuperscript{\rm 1}\quad
Jian Ren\textsuperscript{\rm 2}\quad
Long Zhao\textsuperscript{\rm 3}\quad
Davide Testuggine\textsuperscript{\rm 2}\quad
Xi Peng\textsuperscript{\rm 1}  
\and
\textsuperscript{\rm 1}University of Delaware \quad
\textsuperscript{\rm 2}Snap Inc. \quad
\textsuperscript{\rm 3}Google Research \\ 
{\tt\small \{mengma,xipeng\}@udel.edu}, {\tt\small jren@snap.com}, {\tt\small longzh@google.com}, {\tt\small davide.testuggine@gmail.com}
}

% \author{
% Mengmeng Ma\\
% University of Delaware\\
% {\tt\small mengma@udel.edu}
% \and
% Jian Ren\\
% Snap Inc.\\
% {\tt\small jren@snap.com}
% \and
% Long Zhao\\
% Google Research\\
% {\tt\small longzh@google.com}
% \and
% Davide Testuggine\\
% Snap Inc.\\
% {\tt\small davide.testuggine@gmail.com}
% \and
% Xi Peng \\
% University of Delaware\\
% {\tt\small xipeng@udel.edu}

% }

\maketitle

%%%%%%%%% ABSTRACT
\begin{abstract}

Multimodal data collected from the real world are often imperfect due to missing modalities. Therefore multimodal models that are robust against modal-incomplete data are highly preferred. Recently, Transformer models have shown great success in processing multimodal data. However, existing work has been limited to either architecture designs or pre-training strategies; whether Transformer models are naturally robust against missing-modal data has rarely been investigated. In this paper, we present the first-of-its-kind work to comprehensively investigate the behavior of Transformers in the presence of modal-incomplete data. Unsurprising, we find Transformer models are sensitive to missing modalities while different modal fusion strategies will significantly affect the robustness. What surprised us is that the optimal fusion strategy is dataset dependent even for the same Transformer model; there does not exist a universal strategy that works in general cases. Based on these findings, we propose a principle method to improve the robustness of Transformer models by automatically searching for an optimal fusion strategy regarding input data. Experimental validations on three benchmarks support the superior performance of the proposed method.
\end{abstract}

%%%%%%%%% BODY TEXT
\section{Introduction}
\label{sec:intro}

Multimodal Transformers are emerging as the dominant choice in multimodal learning across various tasks~\cite{dosovitskiy2021an}, including classification~\cite{nagrani2021attention,lee2021parameter}, segmentation~\cite{strudel2021segmenter}, and cross-model retrieval~\cite{kim2021vilt}. They have become the driving force in obtaining better performance on these tasks through a pre-train-and-transfer~\cite{bommasani2021opportunities} paradigm. 

% Multimodal data help the target task by providing additional information, \emph{e.g.}, adding poster images of IMDb movies helps improve a unimodal baseline~\cite{arevalo2017gated}. While constructed datasets have full coverage on every modality, real-world data generally do not. 
Although Transformers have demonstrated remarkable success in processing multimodal data, they generally require modal-complete data. The completeness of modality may not always hold in the real world due to privacy or security constraints. For example, a social network might be unable to access location information if users decline to share their private location~\cite{kossinets2006effects}; a healthcare application might not have all the records available when patients are unwilling to undergo risky or invasive examinations~\cite{suo2019metric}. For this reason, it is important that Transformer models are \textit{robust against missing-modal data}, \textit{i.e.}, the model performance does not degrade dramatically.

% To measure this, we introduce a novel task: we always train the model with full coverage of all modalities, and then we study the test set performance by removing different fractions of each modality. Table \ref{tab:robustness} shows this behavior for a few data points for one particular model, with the model's Macro F1 going lower than a unimodal baseline.

\begin{table}[t]
\caption{Evaluation of the Transformer robustness against missing-modal data on MM-IMDb, UPMC Food-101, and Hateful Memes. We use ViLT~\cite{kim2021vilt} as the backbone. Note that the multimodal performance is \emph{even worse} than the unimodal one, when modality is missing severely (results are highlighted in shaded gray). $^*$The reported evaluation scores are F1-Macro (MM-IMDb), Accuracy (UPMC Food-101), and AUROC (Hateful Memes). Higher scores indicate better results.}\label{tab:robustness}
\centering
\resizebox{\linewidth}{!}{
\begin{tabular}{ccccccc}
\toprule
\multirow{2}{*}{Dataset}                            & \multicolumn{2}{c}{Training} & \multicolumn{2}{c}{Testing}        & \multicolumn{1}{c}{\multirow{2}{*}{Evaluation$^*$}} & \multirow{2}{*}{$\Delta \downarrow$} \\ \cmidrule{2-5}
                                                    & Image         & Text         & Image & Text                       & \multicolumn{1}{c}{}                            &                        \\ \midrule
\multicolumn{1}{c|}{\multirow{3}{*}{MM-IMDb~\cite{arevalo2017gated}}}       & 100\%         & 100\%        & 100\% & \multicolumn{1}{c|}{100\%} & 55.3                                           & 0\%                      \\
\multicolumn{1}{c|}{}                               &   100\%         &  100\%        & 100\% & \multicolumn{1}{c|}{30\%}  & \cellcolor{Gray} 31.2                                           & \textbf{43.6\%}                      \\
\multicolumn{1}{c|}{}                               & 100\%         & 0\%          & 100\% & \multicolumn{1}{c|}{0\%}   & \cellcolor{Gray} 35.0                                           & 36.7\%                      \\ \midrule
\multicolumn{1}{c|}{\multirow{3}{*}{UPMC Food-101~\cite{wang2015upmcfood}}} & 100\%         & 100\%        & 100\% & \multicolumn{1}{c|}{100\%} & 91.9                                           & 0\%                      \\
\multicolumn{1}{c|}{}                               & 100\%         &  100\%        &  100\% & \multicolumn{1}{c|}{30\%}  & \cellcolor{Gray} 65.9                                           & \textbf{28.3\%}                      \\
\multicolumn{1}{c|}{}                               & 100\%         & 0\%          & 100\% & \multicolumn{1}{c|}{0\%}   & \cellcolor{Gray} 71.5                                           & 22.2\%                      \\ \midrule
\multicolumn{1}{c|}{\multirow{3}{*}{Hateful Memes~\cite{douwe2020hatefulmemes}}} & 100\%         & 100\%        & 100\% & \multicolumn{1}{c|}{100\%} & 70.2                                           & 0\%                      \\
\multicolumn{1}{c|}{}                               & 100\%         &  100\%        & 100\% & \multicolumn{1}{c|}{ 30\%}  &  60.2                              & 14.2\%                      \\
\multicolumn{1}{c|}{}                               & 100\%         & 0\%          & 100\% & \multicolumn{1}{c|}{0\%}   &  56.3                                           & \textbf{19.8\%}                      \\ \bottomrule
\end{tabular}}
\end{table}

{Despite its real-world importance, the robustness against missing modalities in multimodal Transformers is seldom investigated in the literature. So far, research on Transformer models has been limited to developing new architectures for fusion~\cite{nagrani2021attention,sun2019videobert,tsai2019multimodal} or exploring better self-supervised learning tasks~\cite{zellers2021merlot,akbari2021vatt,chen2020uniter,zhang2021ufcbert,gabeur2020multi}. Recent work on Transformer robustness has primarily focused on noisy inputs rather than missing modalities~\cite{li2020closer}. } 

A question naturally arises: \textit{Are Transformer models robust against missing-modal data?} We empirically evaluate this problem across multiple datasets in Table~\ref{tab:robustness}. Unsurprisingly, we find that \textit{Transformer models degrade dramatically with missing-modal data.}
As shown, the multimodal performance drops when tested with modal-incomplete data, and, surprisingly, the multimodal performance is even worse than the unimodal when text are missing severely, \textit{i.e.,} only 30\% of text are available.

\begin{table}[t]
\caption{ Evaluation of the Transformer models under different fusion strategies on MM-IMDb and Hateful Memes. \textit{Early} fusion refers to fusion at the first layer; \textit{Late} fusion refers to fusion at the last layer. Different fusion strategies affect model robustness against the missing-modal data.}\label{tab:optiamlfusion}
\centering 
\resizebox{\linewidth}{!}{
\begin{tabular}{ccccccc}
\toprule
\multirow{2}{*}{Dataset}           & \multicolumn{2}{c}{Train} & \multicolumn{2}{c}{Test}           & \multicolumn{2}{c}{Fusion Strategy} \\ \cmidrule{2-7}
                                   & Image       & Text        & Image & Text                       & Early             & Late            \\ \midrule
\multicolumn{1}{c|}{MM-IMDb}       & 100\%       & 100\%       & 100\% & \multicolumn{1}{c|}{100\%} & \textbf{55.3}              & 54.9            \\
\multicolumn{1}{c|}{UPMC Food-101} & 100\%       & 100\%       & 100\% & \multicolumn{1}{c|}{100\%} & \textbf{91.9}              & 91.8            \\
\multicolumn{1}{c|}{Hateful Memes} & 100\%       & 100\%       & 100\% & \multicolumn{1}{c|}{100\%} & \textbf{70.2}              & 64.5            \\ \midrule
\multicolumn{1}{c|}{MM-IMDb}       & 100\%       & 100\%       & 100\% & \multicolumn{1}{c|}{30\%}  & \textbf{31.2}              & 31.0            \\
\multicolumn{1}{c|}{UPMC Food-101} & 100\%       & 100\%       & 100\% & \multicolumn{1}{c|}{30\%}  & 65.9              & \textbf{69.1}            \\
\multicolumn{1}{c|}{Hateful Memes} & 100\%       & 100\%       & 100\% & \multicolumn{1}{c|}{30\%}  & \textbf{60.2}              & 57.8            \\ \bottomrule
\end{tabular}}

\end{table}

% \begin{table}[t]
% \small
% \caption{Performance of Multimodal Transformer under different fusion strategies on MM-IMDb and Hateful Memes. \textit{Early} fusion refers to fusion at the first layer; \textit{Late} fusion refers to fusion at the last layer.}\label{tab:optiamlfusion}
% \centering
% % \resizebox{1\linewidth}{!}{
% \begin{tabular}{ccccc}
% \toprule
% % \multirow{2}{*}{Dataset} & \multicolumn{2}{c}{Test} & \multicolumn{2}{c}{Fusion Strategy} \\ \cmidrule{2-5} 
% Dataset &Image & Text& Early& Late \\ \hline
% \multirow{2}{*}{MM-IMDb} & 100\% & \multicolumn{1}{c|}{100\%} & {55.3} & {54.9} \\
% & 100\% & \multicolumn{1}{c|}{30\%}  & {31.2}&  31.0\\
%  \midrule
% \multirow{2}{*}{Hateful Memes} & 100\% & \multicolumn{1}{c|}{100\%} &\textbf{70.2}& 64.5\\
% & 100\% & \multicolumn{1}{c|}{30\%}  & \textbf{60.2}& 57.8\\
% \bottomrule
% \end{tabular}

% \end{table}
% Another key consideration for multimodal Transformer is the fusion strategy~\cite{bommasani2021opportunities}. 
{Prior work on Transformer models has shown that fusion strategies affect computation complexity and performances~\cite{lee2021parameter,nagrani2021attention,bommasani2021opportunities}.} Another question arises: \textit{Will the fusion strategy affect Transformer robustness against modal-incomplete data?} Unsurprisingly, we observe that different fusion strategies will significantly affect the robustness. What surprised us is that \textit{the optimal fusion strategy is dataset-dependent; there does not exist a universal strategy that works in general cases in the presence of modal-incomplete data.} As shown in Table~\ref{tab:optiamlfusion}, when tested with missing-modal data, early fusion is preferred on MM-IMDb and Hateful Memes, while late fusion is preferred on the UPMC Food-101. This motivates us to improve the robustness of Transformers by automatically attain the optimal fusion strategy regarding different datasets.

We propose a new method to achieve this goal. Our main idea is to jointly optimize Transformer models with modal-complete and modal-incomplete data via multi-task optimization. On top of that, we propose a searching algorithm to attain the best fusion strategy regarding different datasets.
% Furthermore, we learn a policy to inform the Transformer of the best fusion layers to achieve better robustness. Due to the non-differentiable nature of the discrete search space, it is nontrivial to learn such a policy.
% without using policy network~\cite{guo2019spot} or reinforcement learning~\cite{wu2018blockdrop}.
% : without extensive increase model size or time-consuming training process. 
Overall, the main contributions are as follows:
% The main contributions are multi-fold:
\begin{itemize}
    \item To the best of our knowledge, this paper is first-of-its-kind study to investigate the Transformer robustness against modal-incomplete data.
    \item We observe that Transformer models degrade dramatically with missing-modal data. And surprisingly, the optimal fusion strategy is dataset dependent; there does not exist a universal strategy that works in the presence of modal-incomplete data.
    \item We improve the robustness of Transformer models via multi-task optimization. To further improve robustness, we develop an differentiable algorithm to attain the optimal fusion strategy.
    \item We conduct extensive experiments and ablation study on MM-IMDb~\cite{arevalo2017gated}, UPMC Food-101~\cite{wang2015upmcfood}, and Hateful Memes~\cite{douwe2020hatefulmemes} to support our findings and validate the robustness of our method against missing modality.
\end{itemize}

\section{Related Work}

\noindent\textbf{Multimodal learning}.
Different modalities, \textit{e.g.,} natural language, visual signals, or vocal signals, are often complementary in content while overlapping for a common concept. Multimodal learning aims to utilize the complementary information of each modality to improve the performance of various computer vision tasks. A key aspect of multimodal learning is exploring efficient methods for multimodal fusion. Simple methods like concatenation have been widely studied in~\cite{wang2017select,poria2016convolutional}. For efficient cross-modality interaction, a tensor fusion~\cite{zadeh2017tensor} mechanism is proposed by Zadeh \textit{et al.} Following this effort, efficient low-rank fusion~\cite{liu2018efficient} is proposed to address the exponential dimension explosion of tensor fusion. 

The aforementioned fusion mechanisms heavily depend on the completeness of modality, making multimodal fusion impossible with modality-incomplete data. Therefore, another important direction in multimodal learning is to build models that are robust against the modality-incomplete data~\cite{tsai2018learning,ma2021smil}. 
% Tsai \textit{et al.}~\cite{tsai2018learning} propose to learn a generative model and discriminative model during training simultaneously. In this way, when tested with modality-incomplete samples, the generative model can estimate the missing modality for the discriminative model. Instead of generating the original input, 
For example, Ma \textit{et al.}~\cite{ma2021smil} propose a method based on Bayesian Meta-Learning to estimate the latent feature of the modality-incomplete data. However, the existing endeavors usually adopt modality-specific models for each modality, such as ResNet~\cite{he2016deep} for images and LSTM~\cite{hochreiter1997long} for texts, which may lead to a larger set of architectural decisions and training parameters. Instead, we use Transformers as general architectures to jointly model each modality, leading to a simple design and reduced training parameters.

\noindent\textbf{Multimodal transformer}.
Multimodal Transformers have been used in various tasks such as cross-model retrieval~\cite{kim2021vilt,li2021align}, action recognition~\cite{nagrani2021attention}, and image segmentation~\cite{ye2019cross,strudel2021segmenter}. 
They provide several advantages over conventional backbones, \textit{e.g.,} ResNet~\cite{he2016deep}, regarding to flexibility and training load.

The \emph{flexibility} to accommodate modality-incomplete samples is crucial for multimodal backbones, as the real-world multimodal data are often imperfect due to missing modality. Conventional backbones~\cite{pham2019found,tsai2018learning} are generally not flexible. These backbones output the joint multimodal representation by explicitly fusing the features of each modality via concatenation~\cite{poria2016convolutional}, tensor fusion~\cite{zadeh2017tensor}, and others mechanisms. However, explicit fusion requires the presence of all modalities. Missing any modality will break the training pipeline. In contrast, multimodal Transformers use the self-attention mechanism~\cite{vaswani2017attention} to generate a holistic representation of all modalities, allowing the absence of any modalities. When dealing with modality-incomplete samples, it can ignore the absent modalities by applying a mask on the attention matrix. Therefore, multimodal Transformers are more flexible in dealing with missing modalities. Besides, an \emph{easy-to-train} model is vital for multimodal learning. The training load of conventional multimodal backbone grows as the number of modalities increases since the backbone usually consists of modality-specific sub-models that need to be trained independently for each modality~\cite{vielzeuf2018centralnet,ma2021smil}. Instead, Transformer models process all modalities simultaneously using a single model~\cite{kim2021vilt,li2019visualbert}, which greatly reduces the training load. 

% Besides, the pre-trained multimodal Transformer is more \textbf{data efficient} than conventional multimodal backbones. Over the years, pre-trained large-scale multimodal Transformers had been made public available~\cite{li2019visualbert,chen2020uniter,kim2021vilt,zellers2021merlot,tan2021vimpac}. And it has been shown that Transformers~\cite{brown2020fewshot} that are pre-trained on large-scale language datasets are a good few-shot learner. 

% Similarly, we verify the data efficiency of the pre-trained multimodal Transformer on the MM-IMDb dataset~\cite{arevalo2017gated}. Following the exact same setting as SMIL~\cite{ma2021smil}, the finetuned ViLT~\cite{kim2021vilt} outperforms the SMIL~\cite{ma2021smil} by 17\% without relying on the advanced learning paradigm, \textit{e.g.,} Bayesian Meta-Leaning~\cite{lee2020Learning}.

\noindent\textbf{Dynamic neural networks}.
Our work is also related to dynamic neural networks, which adapt the network structure to different inputs, resulting in noticeable gains in accuracy, computation efficiency, or flexibility~\cite{han2021dynamic}. We follow the spirit of the Dynamic Depth~\cite{han2021dynamic} method. Numerous methods have been proposed to dynamically select layers for inference to reduce computation costs. 
Our idea is inspired by AdaShare~\cite{sun2019adashare}, which focuses on learning a policy that selects layers for sharing in multi-task learning. Its main idea is to use Gumbel Softmax Sampling~\cite{jang2016categorical,maddison2016concrete} to learn the policy and network parameters without replying on Reinforcement Learning~\cite{wu2018blockdrop} or extra policy network~\cite{guo2019spot}. However, the direct application of Gumbel Softmax Sampling to our problem results in a large search space with many invalid policies. As a result, we develop an efficient method without using Gumbel Softmax Sampling.

\begin{table}[t]
\caption{ Multi-label classification scores (\%) on the \textit{MM-IMDb}~\cite{arevalo2017gated} under different settings: train and test with full modality (100\% Image + 100\% Text); train and test with single modality (100\% Image or 100\% Text). $^\dagger$ indicate our implementation.}\label{tab:MM-IMDb}
\centering
\resizebox{\linewidth}{!}{
\begin{tabular}{cllcccc}
\toprule
\multirow{2}{*}{Method}                          & \multicolumn{2}{c}{Modality}   & \multirow{2}{*}{F1 Micro} & \multirow{2}{*}{F1 Macro} & \multirow{2}{*}{F1 Weighted} & \multirow{2}{*}{F1 Samples} \\ \cline{2-3}
                                                 & Image & Text                   &                           &                           &                              &                             \\ \midrule
\multicolumn{1}{c|}{\multirow{2}{*}{MFAS~\cite{Perez-Rua_2019_CVPR}}}       & \cmark     & \multicolumn{1}{l|}{}  & 47.8                     & 25.6                     & 42.1                        & 48.4                       \\
\multicolumn{1}{c|}{}                            &       & \multicolumn{1}{l|}{\cmark} & 60.2                     & 48.9                     & 58.5                        & 60.6                       \\ \midrule
\multicolumn{1}{c|}{\multirow{2}{*}{CentralNet~\cite{vielzeuf2018centralnet}}} & \cmark     & \multicolumn{1}{l|}{}  & ---                       & 33.5                    & 49.2                       & ---                         \\
\multicolumn{1}{c|}{}                            &       & \multicolumn{1}{l|}{\cmark} & ---                       & 45.9                    & 57.5                       & ---                         \\ \midrule
\multicolumn{1}{c|}{\multirow{2}{*}{ViLT~\cite{kim2021vilt}$^\dagger$ }}       & \cmark     & \multicolumn{1}{l|}{}  & 51.8                    & 35.0                    & 48.0                       & 51.1                      \\
\multicolumn{1}{c|}{}                            &       & \multicolumn{1}{l|}{\cmark} & 63.3                    & 52.5                    & 62.0                       & 62.9                      \\ \midrule
\multicolumn{1}{c|}{MFAS~\cite{Perez-Rua_2019_CVPR}}                        & \cmark     & \multicolumn{1}{l|}{\cmark} & ---                       & 55.7                    & 62.5                       & ---                         \\
\multicolumn{1}{c|}{CentralNet~\cite{vielzeuf2018centralnet}}                  & \cmark     & \multicolumn{1}{l|}{\cmark} & 63.9                     & 56.1                     & 63.1                        & 63.9                       \\
\multicolumn{1}{c|}{ViLT~\cite{kim2021vilt}$^\dagger$ }                        & \cmark     & \multicolumn{1}{l|}{\cmark} & 64.7                    & 55.3                    & 64.4                       & 64.6                      \\ \bottomrule
\end{tabular}}
\end{table}

\section{Analysis of Multimodal Transformer}

\subsection{Background}

In this paper, we focus on the multimodal Transformer that adopts Vision Transformer (ViT)~\cite{dosovitskiy2021an} as the backbone. ViT consists of a sequence of $L$ Transformer layers, each of which contains a Multi-Head Attention (MHA) layer, Multilayer Perceptron (MLP), and Layer Normalization (LN). The MHA computes the dot-product attention~\cite{vaswani2017attention} on the input sequence, resulting in an attention matrix indicating the similarity between each token.

We follow the vision-language Transformer~\cite{kim2021vilt} to preprocess the data. The input text is mapped into the word embedding through a word embedding codebook and a position embedding codebook. The input image is first partitioned into patches and then flattened into vectors. These vectors are then transformed into latent embedding using linear projection and position embedding. Finally, the image and text embedding are integrated with their corresponding modality-type embedding~\cite{kiela2019supervised,kim2021vilt}. The final multimodal input sequence is the concatenation of vision and text embedding.
% The hierarchy and formulation of the vision-language Transformer is as follow:
% \begin{equation}
% \begin{array}{l}
% {\hat{t}} = [{t_{\text{class}}}; {E_{T}}t_1;\cdots;E_{T}t_{L}] + E^{\text{pos}}_{T}, \\
% {\hat{v}} = [v_{\text{class}};E_{V}v_1;\cdots;E_{V}v_{N}] + E^{\text{pos}}_{V}, \\
% {z_o} = [\hat{t}+t^{\text{type}};\hat{v}+ v^{\text{type}} ],\\
% {\bar{z}_m} = \text{MHA}(\text{LN}(z_{m-1}))+ z_{m-1}, m = 1\cdots M, \\
% {z_m} = \text{MLP}(\text{LN}(\bar{z}_{m}))+ \bar{z}_{m},\;\;\;\;\;\;\;\;  m = 1\cdots M. \\
% \end{array}
% \end{equation}

\subsection{Robustness Against Missing Modality}
\label{subsec3:robust}
\textit{Question: Are Transformer models robust against modal-incomplete data?}

\textit{{Observation:} Unsurprisingly, Transformer models degrade dramatically with modal-incomplete data.}

\begin{table}[t]
\caption{ Classification accuracy (\%) on the \textit{UPMC Food-101}~\cite{wang2015upmcfood}. $^\dagger$ indicate our implementation.}\label{tab:food}
\centering
\resizebox{0.723\linewidth}{!}{
\begin{tabular}{cccc}
\toprule
\multirow{2}{*}{Method}                         & \multicolumn{2}{c}{Modality}       & \multirow{2}{*}{Accuracy} \\ \cline{2-3}
                                                & Image & Text                       &                           \\ \midrule
\multicolumn{1}{c|}{\multirow{2}{*}{BERT+LSTM~\cite{ignazio2020bertlstm}}} & \cmark & \multicolumn{1}{c|}{}      & 71.7                     \\
\multicolumn{1}{c|}{}                           &       & \multicolumn{1}{c|}{\cmark} & 84.4                     \\ \midrule
\multicolumn{1}{c|}{\multirow{2}{*}{ViLT~\cite{kim2021vilt}$^\dagger$}}      & \cmark & \multicolumn{1}{c|}{}      & 71.5                     \\
\multicolumn{1}{c|}{}                           &       & \multicolumn{1}{c|}{\cmark} & 84.4                     \\ \midrule
\multicolumn{1}{c|}{BERT+LSTM~\cite{ignazio2020bertlstm}}                  & \cmark & \multicolumn{1}{c|}{\cmark} & 92.5                     \\
\multicolumn{1}{c|}{MMBT~\cite{kiela2019supervised}}                       & \cmark & \multicolumn{1}{c|}{\cmark} & 92.1                     \\
\multicolumn{1}{c|}{ViLT~\cite{kim2021vilt}$^\dagger$}                       & \cmark & \multicolumn{1}{c|}{\cmark} & 92.0                     \\ \bottomrule
\end{tabular}}

\end{table}

We begin by defining how Transformer robustness is measured. Specifically, we adopt two different evaluation settings: a ``full'' test set with full-modal data and a ``missing'' test set with missing-modal data. We evaluate Transformer robustness by comparing model performance on the ``missing" test set to the ``full" test set: the smaller the difference, the better the robustness.

% A robust model capable of making correct predictions in the absence of some missing modalities is critical for the real-world application of multimodal learning.  We use the ViLT~\cite{kim2021vilt} as our backbone to conduct evaluation.
% How robust is multimodal Transformer to missing modality? 
First, we empirically verify that model performance degrades dramatically in the presence of missing-modal data. Table~\ref{tab:robustness} shows the evaluation results on three widely-used multimodal datasets. As shown, when only 30\% text modality is observed, the multimodal performance drops by 43.6\%, 28.3\%, and 14.2\%, respectively. Moreover, when the modality is severely missing, the multimodal performance is even worse than the unimodal one on MM-IMDb and UPMC Food-101.

% Moreover, the model's performance dropped more on MM-IMDb and UPMC Food-101. 
Second, we observe that the modality importance varies on different datasets. We use unimodal performance to indicate the importance of each modality. Results of unimodal performance are shown in Tables~\ref{tab:MM-IMDb}, \ref{tab:food}, and \ref{tab:hategulmemes}. As shown, the text modality is more important than the image modality on MM-IMDb and Food-101, while text and image are equally important on Hateful Memes. In specific, in the first two datasets, text has higher performance than the image. Moreover, the performance gap between unimodal (text) and full modalities is smaller then that between unimodal (image) and full modalities ($10\%$ vs.~$22\%$), indicating that text is the dominant modality. In contrast, in the Hateful Memes dataset, the performances of text and image are comparable, and the performance gap between multimodal and unimodal is large ($>$20\%), demonstrating that the two modalities are equally important.
% For a dataset in which the importance of each modality is distinct, we refer to the dominant modality as the strong modality and the supportive modality as the weak modality. Therefore, we can conclude that \textit{multimodal Transformers are sensitive to modality-incomplete data, especially for missing the strong modalities.}

\begin{table}[t]
\caption{ AUROC (\%) on the unseen test set of \textit{Hateful Memes}~\cite{douwe2020hatefulmemes}. $^*$denotes the results from  hateful memes challenge~\cite{kiela21report}. $^\dagger$ indicate our implementation. }\label{tab:hategulmemes}
\centering
\resizebox{0.7\linewidth}{!}{
\begin{tabular}{cccc}
\toprule
\multirow{2}{*}{Method}                              & \multicolumn{2}{c}{Modality}                                       & \multirow{2}{*}{AUROC} \\ \cline{2-3}
                                                     & Image                 & Text                                       &                               \\ \midrule
\multicolumn{1}{c|}{\multirow{2}{*}{Unimodal$^*$}}       & \cmark & \multicolumn{1}{c|}{}                      & 54.6                         \\
\multicolumn{1}{c|}{}                                &                       & \multicolumn{1}{c|}{\cmark} & 62.7                         \\ \midrule
\multicolumn{1}{c|}{\multirow{2}{*}{ViLT~\cite{kim2021vilt}$^\dagger$}}           & \cmark & \multicolumn{1}{c|}{}                      & 56.3                         \\
\multicolumn{1}{c|}{}                                &                       & \multicolumn{1}{c|}{\cmark} & 58.3                         \\ \midrule
\multicolumn{1}{c|}{MMBT-Grid~\cite{kiela2019supervised}$^*$}                     & \cmark & \multicolumn{1}{c|}{\cmark} & 67.3                         \\
\multicolumn{1}{c|}{MMBT-Region~\cite{kiela2019supervised}$^*$}                     & \cmark & \multicolumn{1}{c|}{\cmark} & 72.2                         \\
\multicolumn{1}{c|}{ViLBERT~\cite{jiasen2019vilbert}$^*$}                         & \cmark & \multicolumn{1}{c|}{\cmark} & 73.4                         \\
\multicolumn{1}{c|}{ViLBERT CC~\cite{jiasen2019vilbert}$^*$}                      & \cmark & \multicolumn{1}{c|}{\cmark} & 72.8                         \\
\multicolumn{1}{c|}{Visual BERT~\cite{li2019visualbert}$^*$}                         & \cmark & \multicolumn{1}{c|}{\cmark} & 73.2                         \\
\multicolumn{1}{c|}{ViLT~\cite{kim2021vilt}$^\dagger$} & \cmark & \multicolumn{1}{c|}{\cmark} & 70.2                         \\ \bottomrule
\end{tabular}}
\end{table}

Finally, we empirically observe that Transformer models tend to overfit to dominate modalities. Specifically, we first train our model with multimodal data and test with different unimodal data. Then we examine the performance gap between unimodal and multimodal testing -- the larger the gap, the more severe the overfitting. Experimental results are shown in Table~\ref{tab:hypothesis}. As shown, for MM-IMDb dataset, text-only testing performs better than image-only testing, which means that text-only testing is closer to full-modal testing. Therefore text-only testing has a smaller gap than image-only testing, indicating that models trained on this dataset tend to overfit to text modality. 
% Recall that text modality is the dominant modality on MM-IMDb.
% Therefore, missing strong modalities will severely degrade model's performance.

\subsection{Optimal Fusion Strategy}
\label{subsec3:layer}
\textit{Question: Will the fusion strategy affect Transformer robustness against modal-incomplete data?}

\textit{Observation: Different fusion strategies do affect the robustness of Transformer models. Surprisingly, the optimal fusion strategy is dataset-dependent; there does not exist a universal strategy that works in general cases.}

Typically, there exists two widely used fusion strategies: early and late fusion. For early fusion, cross-modal interaction happened in early layers, ensuring the model to have sufficient capacity to exploit multimodal information, but at the expense of larger computing costs. For late fusion, cross-modal interaction happened in later layers, which significantly lower computation costs, but the resulting model might have limited capacity to take the full advantage of multimodal information. 

%  It remains an open question that how to determine the optimal layer for multimodal fusion~\cite{bommasani2021opportunities}. Existing solution on multimodal Transformer adopt a fixed fusion strategy. For example, VisualBERT~\cite{li2019visualbert} and ViLT~\cite{kim2021vilt} adopt early fusion; ViLBERT~\cite{jiasen2019vilbert} and AVBERT~\cite{lee2021parameter} allow cross-modal interaction after the middle layer; CLIP~\cite{radford2021clip} conducts multimodal interaction at the last layer. However, the one-size-fits-all approach may not be optimal on all datasets. As shown in Table~\ref{tab:optiamlfusion}, when tested with missing-modal data, early fusion is preferred on MM-IMDb and Hateful Memes, while late fusion is preferred on the UPMC Food-101.
 It remains an open question on how to determine the optimal layer for fusion~\cite{bommasani2021opportunities}. Existing solutions in multimodal Transformer adopt a fixed fusion strategy~\cite{li2019visualbert,kim2021vilt,jiasen2019vilbert,lee2021parameter,radford2021clip}. However, the one-size-fits-all approach may not be optimal on all datasets. As discussed in Sec.~\ref{sec:intro}, the optimal fusion strategy is dataset-dependent.

\begin{table}[t]
\caption{ Evaluation on the overfitting issue of Transformer models on MM-IMDb and Hateful Memes. Transformer models tend to overfit to dominate modality.}\label{tab:hypothesis}
\centering
\resizebox{0.9\linewidth}{!}{
\begin{tabular}{cccccc}
\hline
\multirow{2}{*}{Dataset}                           & \multicolumn{2}{c}{Training} & \multicolumn{2}{c}{Testing}        & \multirow{2}{*}{Evaluation} \\
                                                   & Image         & Text         & Image & Text                       &                             \\ \hline
\multicolumn{1}{c|}{\multirow{3}{*}{MM-IMDb}}      & 100\%         & 100\%        & 100\% & \multicolumn{1}{c|}{100\%} & 55.3                        \\
\multicolumn{1}{c|}{}                              & 100\%         & 100\%        & 0\%   & \multicolumn{1}{c|}{100\%} & \cellcolor{Gray} 47.4                        \\
\multicolumn{1}{c|}{}                              & 100\%         & 100\%        & 100\% & \multicolumn{1}{c|}{0\%}   & \cellcolor{Gray} 35.0                        \\ \hline
\multicolumn{1}{c|}{\multirow{3}{*}{Hatful Memes}} & 100\%         & 100\%        & 100\% & \multicolumn{1}{c|}{100\%} & 70.2                        \\
\multicolumn{1}{c|}{}                              & 100\%         & 100\%        & 0\%   & \multicolumn{1}{c|}{100\%} & \cellcolor{Gray}55.7                        \\
\multicolumn{1}{c|}{}                              & 100\%         & 100\%        & 100\% & \multicolumn{1}{c|}{0\%}   & \cellcolor{Gray} 54.9                        \\ \hline
\end{tabular}}

\end{table}
\section{Robust Multimodal Transformer}
\label{sec4:method}

\begin{figure*}[t]
    \centering
    \includegraphics[width=0.9\linewidth]{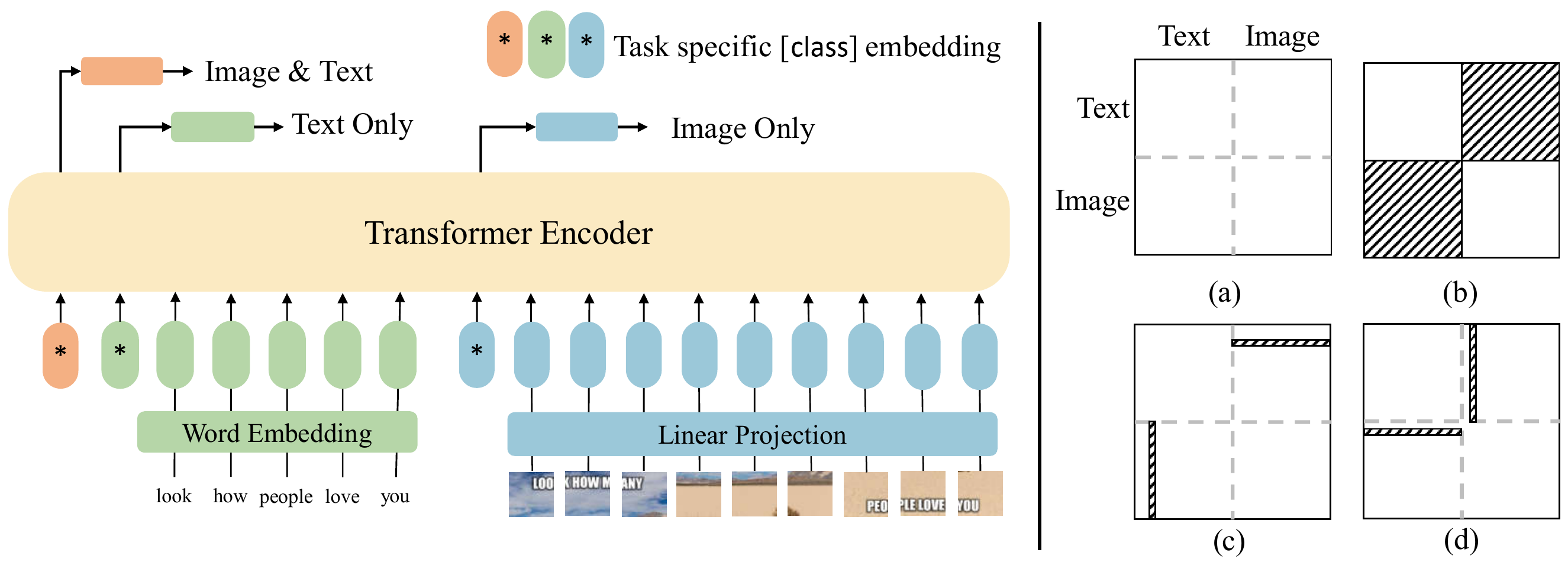} 
    \caption{\emph{Left}: Overview of our model. \emph{Right}: Attention masks for different tasks: (a) Original attention without masking; (b) Mask-out cross-modal attention; (c) Mask-out image attention for text only [class] token; (d) Mask-out text attention for image only [class] token.}
    \label{fig:method}
\end{figure*}
% Our goal is to improve the robustness of multimodal Transformer against missing modality. We propose to tackle . In the following sections, we first describe the design of a robust multimodal Transformer. Then we present how to search for the optimal layer for multimodal fusion.
Without loss of generality, we consider a multimodal dataset with two modalities. Formally, let $\mathcal{D}=\left \{  \bm x^{1}_{i}, \bm x^{2}_{i}, y_{i} \right \}_{i} $ denote the multimodal dataset, where $\bm x^{1}_{i}$ and $\bm x^{2}_{i}$ represent two different modalities and $y_{i}$ is the corresponding label. Our target is to improve the Transformer robustness against modal-incomplete data, \textit{i.e.,} model performance does not degrade dramatically. To this end, we propose to leverage multi-task optimization and the optimal fusion strategy to improve the robustness.

\textbf{Multi-task learning.} We intend to improve the performance of Transformer models in dealing with modal-incomplete data. In the missing-modal scenario, training data are modal-complete, while testing samples are modal-incomplete. This discrepancy motivates us to incorporate missing-modal data in the training process. By doing so, the Transformer model will be more confident in its prediction on modal-incomplete data, resulting in a robust Transformer. The key idea is to leverage the masking mechanism to ``generate" the modal-incomplete data during training and jointly optimize the Transformer model with modal-complete and modal-incomplete data via multi-task optimization. Our method is simple to implement with minimal modification to the Transformer.
% The completeness of modality is different between training and testing samples in the presence of modal-incomplete data. This motivates us to incorporate missing-modal data in the training process, leading to a more robust model. By doing so, the Transformer models might not treat the missing-modal test data as out-of-distribution samples, resulting in more confident prediction. Our key idea is to leverage the masking mechanism to ``generate" the modal-incomplete data during training and jointly optimize the Transformer model with modal-complete and modal-incomplete data via multi-tasking optimization. Our method is simple to implement with minimal modification to the Transformer.

\textbf{Optimal fusion strategy.} The goal is to automatically search for the optimal fusion strategy on different datasets. Manually finding the optimal strategy is not practical, especially for lager-scale models~\cite{brown2020language,zellers2021merlot,dosovitskiy2021an}, due to the heavy training load. However, designing such an algorithm is nontrivial in light of the non-differentiable nature of the discrete searching space~\cite{sun2019adashare}. Existing methods, such as Reinforcement Learning (RL)~\cite{wu2018blockdrop} and policy network (PN)~\cite{guo2019spot}, are either inefficient in training or adding additional parameters to the model. 
% For example, RL is notoriously for the slow training process, and PN will bring extra parameters to the model. 
We propose a differentiable method to obtain the fusion strategy through standard back-propagation. The key idea is to learn a policy to obtain optimal layers for fusion. Specifically, each layer is assigned with a policy parameter\footnote{Policy parameters are negligible compared to model parameters.} to decide fusion or not. The fusion strategy is sampled from the policy parameters.

\subsection{Improve Robustness via Multi-task Learning}
\label{subsec4:robust}

% A robust Transformer can make accurate predictions even with partial modalities. However, as shown in Table \ref{tab:robustness}, the existing multimodal Transformer is not trustworthy when partial modalities are observed. Additionally, when missing strong modalities, the model's multimodal performance is even worse than its unimodal performance. We propose a multi-task learning-based method to improve the robustness of the multimodal Transformer. The idea is to use various purposefully designed tasks to model the full and partial information of the multimodal data. In this way, our model is flexible to use different representations to deal with any type of input data, \textit{i.e.,} full or partial modalities, resulting in a robust model with multimodal performance no worse than the unimodal performance.

% Without loss of generality, we consider a multimodal dataset containing two modalities. Formally, we let $\mathcal{D}=\left \{  \bm x^{1}_{i}, \bm x^{2}_{i}, y_{i} \right \}_{i} $ denote the multimodal dataset, where $\bm x^{1}_{i}$ and $\bm x^{2}_{i}$ represent two different modalities and $y_{i}$ is the corresponding label.
On a bimodal dataset, \textit{e.g.,} image and text, multi-task learning can have up to three distinct tasks: full-modal (image + text) task, image-only task, and text-only task. Let $f_{\theta}$ denote a Transformer parameterized by $\bm \theta$. The total loss function is defined as follows:
% \begin{equation}
% \mathcal{L}_{multi}  = \lambda_1 \mathcal{L}_{img}(\bm x^{1}; \bm \theta) + \lambda_2 \mathcal{L}_{txt}(\bm x^{2}; \bm \theta) + \lambda_3 \mathcal{L}_{it}(\bm x^{1}, \bm x^{2}; \bm \theta),
% \end{equation}
\begin{equation}
\resizebox{0.9\linewidth}{!}{
    $\mathcal{L}  = \lambda_1 \mathcal{L}_{img}(\bm x^{1}; \bm \theta) + \lambda_2 \mathcal{L}_{txt}(\bm x^{2}; \bm \theta) + \lambda_3 \mathcal{L}_{it}(\bm x^{1}, \bm x^{2}; \bm \theta)$,}
\end{equation}
where $\mathcal{L}_{img}$ is the loss for image only task; $\mathcal{L}_{txt}$ is the loss for text only task; $\mathcal{L}_{it}$ is the loss for image + text task; $\lambda_1$, $\lambda_2$, and $\lambda_2$ are hyperparameters to balance each loss.

Transformer models leverage classification tokens~\cite{kim2021vilt,jiasen2019vilbert,vaswani2017attention} to generate embeddings for classification. For the three tasks, we add three classification tokens to the Transformer model. Each classification token will output task-specific embedding for the target task. The model overview is shown in Figure \ref{fig:method} Left. For multi-task learning, each task is expected to use only corresponding modalities for classification, \textit{e.g.}, text modality for the text-only task. Therefore we apply masks on the attention matrix, ensuring that the output embedding of each classification token contains only information from corresponding modalities. For instance, in the text-only task, we mask out all the self-attention to the image and the cross-attention between image and text. The attention masks are shown in Figure \ref{fig:method} Right.

% We introduce three classification tokens for each prediction task. Figure \ref{fig:method} \textbf{Left} shows the overview of our method. Different tasks require the model to leverage different combination of modalities. We apply masks on the attention to ensure each task token only see the task-specific modalities. Visualizations of different attention masks are shown in Figure \ref{fig:method} \textbf{Right}: (a) an original attention matrix without mask. We observe that the attention matrix can be divided into four parts: image to image attention, image to text attention, text to image attention, and text to text attention; (b) a mask that mask-out all cross-modality attention; (c) a mask that mask-out image attention for the text-only [class] token; (d) a mask that mask-out text attention for the image-only [class] token.

\begin{algorithm}[t]
	\caption{ Search for Optimal Fusion Policy.}
	\LinesNumbered
	\label{alg:train}
	\KwIn{Multimodal dataset $D^{tr},D^{val}$; inner-level learning rate $\gamma$; outer-level learning rate $\beta$; initialized policy parameter $\bm \alpha$; number of iterations $K$.}
	\BlankLine
	\While{not converged}{
	    $ \{\mathbf{x}_i^1,\mathbf{x}_i^2, y_i \} \sim D^{tr}; \{\mathbf{x}_j^1, \mathbf{x}_j^2, y_j \} \sim D^{val}$ \\
	    $\bm \theta_{0} \leftarrow \bm \theta$\\
	    \textbf{Lower-Level}: \\
	    \For{$k=0\; \mathrm{to}\; K-1$}{
	    Sample policy $\bm s$ with $\bm \alpha$ using Eqn. \ref{eqn:sampling}\\
	    $\bm \theta_{k+1} \leftarrow \bm \theta_{k} - \gamma \nabla_{\bm \theta_{k}} \mathcal{L}^{tr}(\mathbf{x}_i^1,\mathbf{x}_i^2, \bm s;{\bm \theta})$ 
	    }
        $\bm \theta^{*} \leftarrow \bm \theta_{K}$  \\
	    \textbf{Upper-Level:} \\
	    Sample policy $\bm s$ with $\bm \alpha$ using Eqn. \ref{eqn:sampling}\\
	    $\bm \alpha \leftarrow \bm \alpha - \beta \nabla_{\bm \alpha} \mathcal{L}^{val}(\mathbf{x}_j^1, \mathbf{x}_j^2, \bm s; {\bm \theta^{*}})$  \\
	}
\end{algorithm}

\subsection{Search for the Optimal Fusion Strategy}
\label{subsec4:search}
% Finding the optimal fusion layer of different datasets is critical for multimodal Transformer, leading to saved computation cost, improved robustness, or superior performance. The goal of optimal fusion layer search is to learn a policy that decides whether or not to conduct fusion at each layer. However, learning such a policy is nontrivial as the search space is discrete and non-differentiable. Existing methods usually rely on reinforcement learning~\cite{wu2018blockdrop} or a policy network~\cite{guo2019spot} to learn the policy, resulting in increased model parameters and training complexity. We propose a simple method to jointly learn the optimal policy and network parameters through standard back-propagation. The idea is to formulate the policy learning and network training into a bi-level optimization problem~\cite{liu2018darts}. 
We first introduce the formulation of the search problem. Let $\bm \alpha = \left\{ \alpha_{m} \right\}^{M}_{m=1}$ denote the policy parameters, where $M$ is the total number of layers. To learn the optimal policy parameters, we formulate the parameter learning into a bi-level optimization problem. The object of optimization is to minimize the loss on a validation set $\mathcal{L}^{val}(\bm \alpha, \bm \theta^*)$, where the optimal weights $\bm \theta^*$ are obtained by minimizing the the training loss $\mathcal{L}^{tr}(\bm \theta, \bm \alpha^* )$. The optimization problem is formulated as follows:
\begin{equation}
    \begin{array}{l}
     \underset{\bm \alpha}{\text{min}}\;\; \mathcal{L}^{val}({\bm \theta^*, \bm \alpha}), \\
     \textit{ s.t.}\;\;\; \bm \theta^* = \underset{\bm \theta}{\text{argmin}} \,\mathcal{L}^{tr}(\bm \theta, \bm \alpha^*).
    \end{array}
\end{equation}
% One important step for this optimization problem is to samples the fusion strategy with the policy parameters. 
Next, we describe how to generate the fusion strategy using policy parameters. Existing work on policy learning generally assumes $\alpha_m$ to be bivariate~\cite{sun2019adashare}, resulting in a search space with $2^{M}$ possible policies\footnote{Two actions, fusion or not fusion, for each layer and $M$ layers in total.}. However, the search space can be significantly reduced. In multimodal fusion, one usually conducts fusion starting from a certain layer until the last layer. Following this convention, we set $\alpha_m$ to be univariate, leading to a search space with $M$ policies. Let $\bm Q$, a lower triangular matrix with all non-zero elements equal to one and with the size of $M\times M$, denote all the policies. Let $\bm s$ denote the final policy. To obtain a policy, we first softmax the policy parameter to get a soft policy: $\bm s_s = \text{softmax}(\bm \alpha)$. Then, we convert $\bm s_s$ into a hard policy using one-hot encoding with the differential trick\footnote{ Differentiable trick: $\bm s_h=\text{onehot}(\bm s_s) - \bm s_s.\text{detach()} + \bm s_s$.}: $\bm s_h=\text{onehot}(\bm s_s)$. The final policy can be obtained by sampling from $\bm Q$ using the hard policy $\bm s_h$. 

\begin{equation}
\label{eqn:sampling}
\bm s=\left \langle \bm Q, \bm s_h  \right \rangle.
\end{equation}
Our method significantly reduces the search space, resulting in an differentiable and easy-to-train policy learning process. The overall method is shown in Algorithm \ref{alg:train}. 
Once the optimal policy is learned, we fixed the policy to retrain the model $\bm \theta$ using the whole training set.

% \begin{algorithm}[t]
% 	\caption{Retraining with the Optimal Policy.}\label{alg:retrain}
% 	\LinesNumbered
% 	\KwIn{Multimodal dataset $\mathcal{D}$; learned optimal policy parameter $\bm \alpha^{*}$.}
% 	\BlankLine
% 	\While{not converged}{
% 	    $ \{\mathbf{x}_i^1,\mathbf{x}_i^2, y_i \} \sim \mathcal{D}.$ \\
% 	    \textbf{Network Update}: \\
% 	    Sample policy $\bm s^*$ with $\bm \alpha^{*}$ using Eqn. \ref{eqn:sampling}\\
% 	    $\bm \theta \leftarrow \bm \theta - \beta \nabla_{\bm \theta_{k}} \mathcal{L}_{train}(\mathbf{x}_i^1,\mathbf{x}_i^2,\bm s^*;{\bm \theta})$ 
% 	}
% \end{algorithm}

\section{Experiments}

In this section, we analyze the performance of our approach on three multimodal datasets and aim to answer the following questions: (1) Does the Transformer model perform well with modal-complete data? (Sec. \ref{sec5:welltrain}) (2) Does the proposed method improve the robustness of backbone against missing-modal data? (Sec. \ref{sec5:robust}) (3) Why different datasets prefer different layers for multimodal fusion? (Sec. \ref{sec5:layer}) (4) What factors affect the effectiveness of our method? (Sec. \ref{sec5:ablation})

\subsection{Datasets and Metrics}
\textbf{Datasets.} \textit{MM-IMDb}~\cite{arevalo2017gated } has two modalities: image and text. The target task is to predict the genres of a movie using image, text, or both. This task is multi-label classification, as each movie might have multiple genres. This dataset contains $25,956$ image-text pair and $23$ classes.

\textit{UPMC Food-101}~\cite{wang2015upmcfood} is a classification dataset composed of text and images. The UPMC Food-101 categories are identical to one of the largest publicly available food image datasets: the ETHZ Food-101~\cite{bossard14}. In the UPMC Food-101, image and text pairs are noisy since all the images are obtained in an uncontrolled environment. This dataset contains 90,704 image-text pairs and 101 classes.

\textit{Hateful Memes}~\cite{douwe2020hatefulmemes} is another challenging multimodal dataset that focuses on identifying hate speech in memes. It is constructed to fail models that rely on single modality and multimodal models are likely to perform well: challenging samples (“benign confounders”) are added to the dataset to make relying on unimodal signals more difficult. Hateful Memes contain exactly 10k memes.

\textbf{Metrics.} For \textit{MM-IMDb} dataset, following previous works~\cite{ma2021smil,vielzeuf2018centralnet,kiela2019supervised}, we use F1 Micro, F1 Macro, F1 Samples, and F1 Weighted to evaluate multi-label classification. For \textit{UPMC Food-101}, similar to previous works~\cite{ignazio2020bertlstm,wang2015upmcfood}, we compute the classification accuracy. For \textit{Hateful Memes}, following~\cite{douwe2020hatefulmemes}, we use Area Under the Receiver Operating Characteristic Curve (AUROC) to evaluate model performance.

% \begin{figure*}[t]
%     \centering
%     \includegraphics[width=0.95\linewidth]{figures/fig_robust.pdf} 
%     \caption{Robustness of multimodal Transformer~\cite{kim2021vilt} on MM-IMDb~\cite{arevalo2017gated} (\emph{left}), UPMC Food-101~\cite{wang2015upmcfood} (\emph{middle}), and hateful Memes dataset~\cite{douwe2020hatefulmemes} (\emph{right}). \emph{Our method significantly improves the model's robustness, especially when the modality is severely missing.} Models are trained with 100\% text + 100\% image and tested  with $\eta$\% text + 100\% image. ``Image only'' refers to the single modality setting -- only image modality is used for training and testing. We adopt ViLT~\cite{kim2021vilt} as the baseline model. Reported values are averaged over $10$ runs.}
%     \label{fig:imdb}
% \end{figure*}

\begin{figure*}[t]
\centering
\begin{subfigure}{.32\textwidth}
  \centering
  \includegraphics[width=\linewidth]{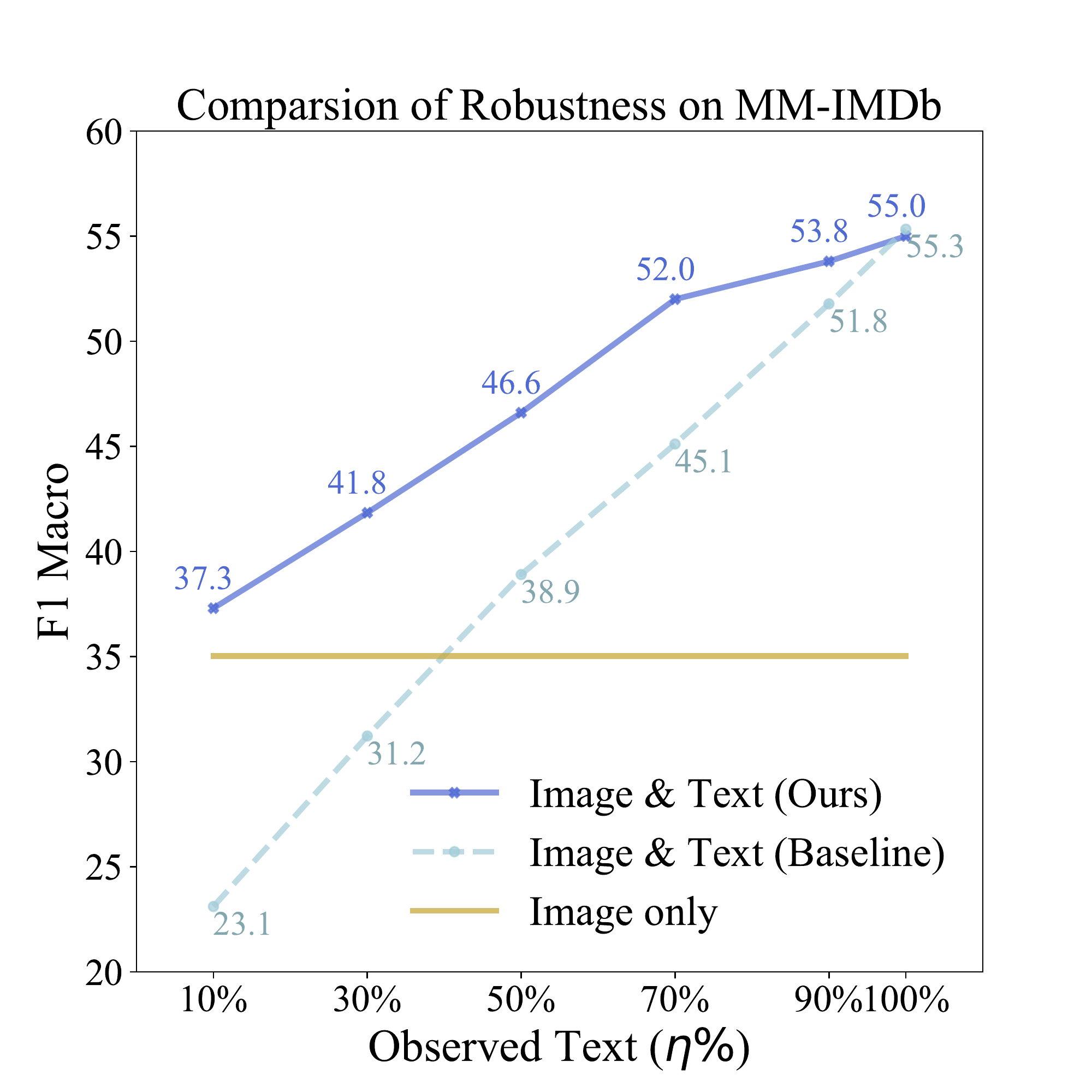}  
\end{subfigure}
\hspace{-0.5em}
\begin{subfigure}{.32\textwidth}
  \centering
  \includegraphics[width=\linewidth]{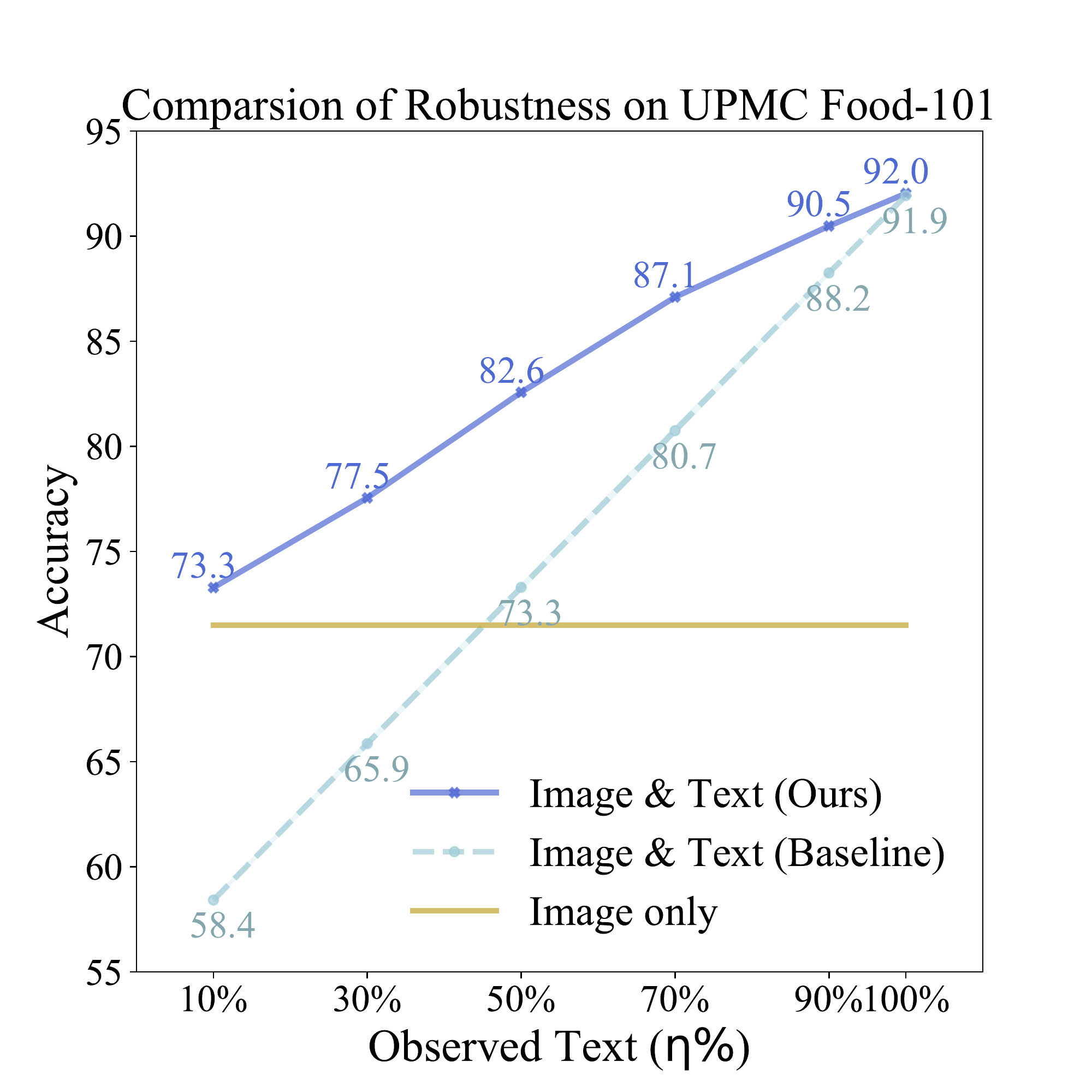}  
\end{subfigure}
\hspace{0.1em}
\begin{subfigure}{.32\textwidth}
  \centering
  \includegraphics[width=\linewidth]{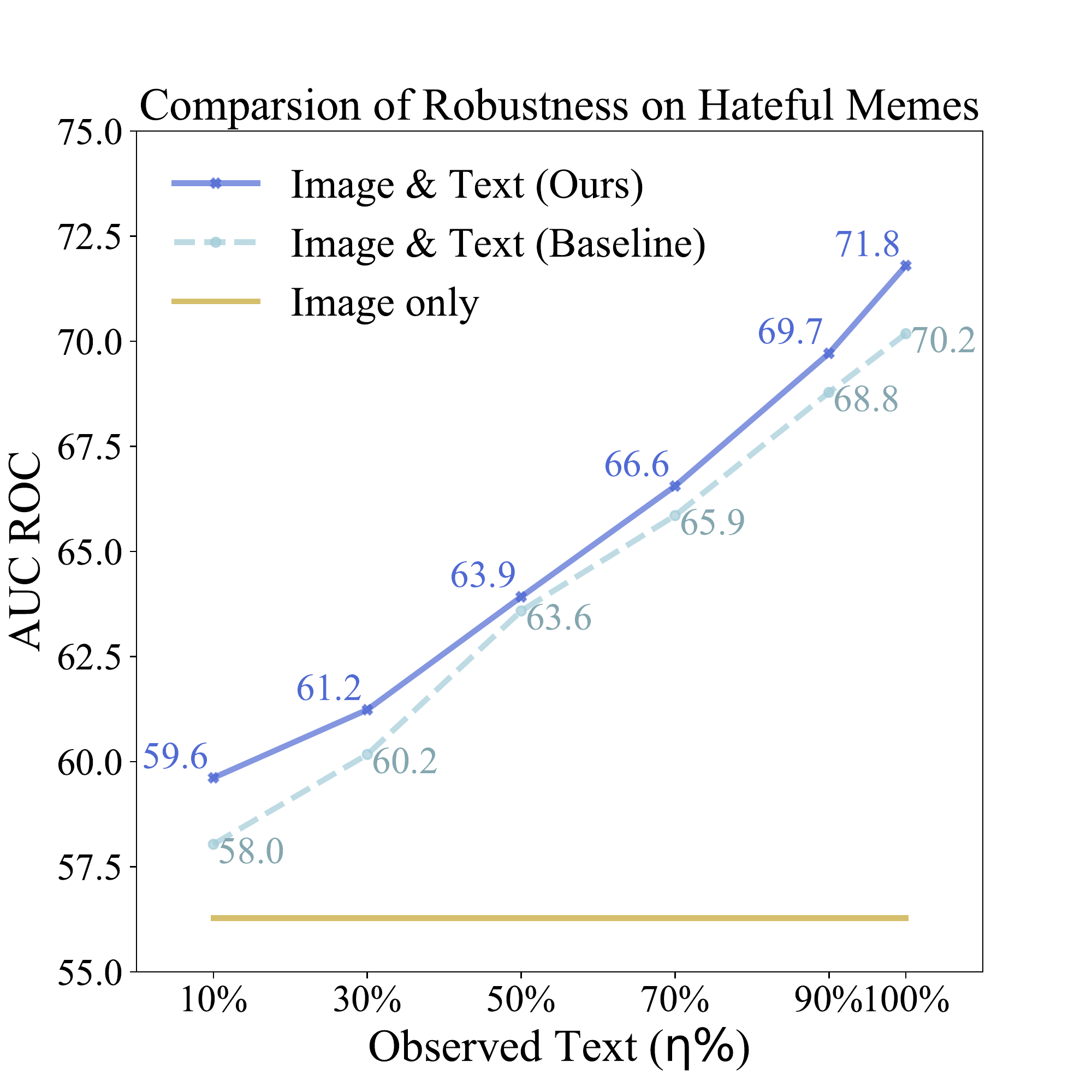}  
\end{subfigure}
\caption{Comparison of Transformer robustness~\cite{kim2021vilt} on MM-IMDb~\cite{arevalo2017gated} (\emph{left}), UPMC Food-101~\cite{wang2015upmcfood} (\emph{middle}), and Hateful Memes dataset~\cite{douwe2020hatefulmemes} (\emph{right}). We adopt ViLT~\cite{kim2021vilt} as the backbone. Models are trained with 100\% text + 100\% image and tested  with $\eta$\% text + 100\% image. ``Image only'' refers to the single modality setting -- only image modality is used for training and testing. \emph{Our method significantly improves model robustness, especially when the modality is severely missing.}}
\label{fig:imdb}
\end{figure*}

% \begin{figure*}[t]
% \centering
% \minipage{0.32\textwidth}
% \centering
%   \includegraphics[width=\linewidth]{figures/imdb.pdf}
% \endminipage
% \hspace{-0.5em}
% \minipage{0.32\textwidth}
% \centering
%   \includegraphics[width=\linewidth]{figures/food_101.pdf}
% \endminipage
% \hspace{-0.2em}
% \minipage{0.32\textwidth}%
% \centering
%   \includegraphics[width=\linewidth]{figures/hatememes.pdf}
% \endminipage
% \caption{Robustness of multimodal Transformer~\cite{kim2021vilt} on MM-IMDb~\cite{arevalo2017gated} (\emph{left}), UPMC Food-101~\cite{wang2015upmcfood} (\emph{middle}), and hateful Memes dataset~\cite{douwe2020hatefulmemes} (\emph{right}). \emph{Our method significantly improves the model's robustness, especially when the modality is severely missing.} Models are trained with 100\% text + 100\% image and tested  with $\eta$\% text + 100\% image. ``Image only'' refers to the single modality setting -- only image modality is used for training and testing. We adopt ViLT~\cite{kim2021vilt} as the baseline model. Reported values are averaged over $10$ runs.}
% \label{fig:imdb}
% \end{figure*}

\subsection{Implementation Details}

\textbf{Multimodal backbone.} We use ViLT as the backbone since it represents the common design of multimodal transformer. ViLT~\cite{kim2021vilt} is a pure Transformer-based model that does not rely on modality-specific sub-models to extract features, and multiple objectives are used to pre-train the model, \textit{e.g.,} Image Text Matching (ITM) and Masked Language Modeling (MLM).

\textbf{Inputs.} For image modality, we resize the input image into $384 \times 384$. Following~\cite{dosovitskiy2021an}, we extract $32 \times 32$ patches from the input images, yielding a total of $12 \times 12 = 144$ patches per image. For text modality, we adopt \textit{bert-base-uncased} tokenizer to tokenize text inputs. The maximum length of the text sequences is various across different datasets: $1024$ (MM-IMDb), $512$ (Food-101), and $128$ (Hateful Memes).

\textbf{Network training.}
We use Adam optimizer~\cite{kingma2015adam} in all experiments with different learning rates for network training and policy learning. For network training, the base learning rate is $3\times 10^{-5}$ and weight decay is $2\times 10^{-2}$. For policy learning, the base learning rate is $3\times 10^{-3}$ and weight decay is $3\times 10^{-5}$. Model parameters are initialized using the pre-trained weights provided by ViLT~\cite{kim2021vilt}.

\subsection{Performance on the Full Test Set}
\label{sec5:welltrain}

We compare our model with other baseline models under the ``full'' evaluation setting, where all modalities are observed.

In Tables \ref{tab:MM-IMDb}, \ref{tab:food}, and \ref{tab:hategulmemes}, we report results on MM-IMDb, UPMC Food-101, and Hateful Memes, respectively. Our results are either state-of-the-art or on par with other methods: on MM-IMDb, compared to CentralNet, we achieve an F1 Weighted of $64.4$ in comparison to $63.1$; on UPMC Food-101, compared to MMBT, the most similar model to ours, we achieve an accuracy of $92.0$ in comparison to $92.1$; on Hateful Memes, we
achieve comparable performance to MMBT ($70.2$ vs. $72.2$). The results demonstrate the superiority of our model. 
% Therefore, the findings based on its results are reliable.

\subsection{Performance on the Missing Test Set}
\label{sec5:robust}

% \begin{figure*}[t]
%     \centering
%     \includegraphics[width=\linewidth]{figures/fig_robust.pdf} 
%     \caption{Robustness of ViLT~\cite{kim2021vilt} on MM-IMDb~\cite{arevalo2017gated} dataset. Training with 100\% text + 100\% image; Testing with $\eta$\% text + 100\% image. Image only refers to the single modality setting (training and testing using image modality only). Reported values are averaged over 10 runs.}
%     \label{fig:imdb}
% \end{figure*}

% \begin{figure}[t]
%     \centering
%     \includegraphics[width=\linewidth]{figures/fig_2.pdf} 
%     \caption{Robustness of ViLT~\cite{kim2021vilt} on UPMC Food-101~\cite{wang2015upmcfood} dataset. Training with 100\% text + 100\% image; Testing with $\eta$\% text + 100\% image. Image only refers to the single modality setting (training and testing using image modality only). Reported values are averaged over 10 runs.}
%     \label{fig:food}
% \end{figure}

% \begin{figure}[t]
%     \centering
%     \includegraphics[width=\linewidth]{figures/fig_hatefulmemes.pdf} 
%     \caption{Robustness of ViLT~\cite{kim2021vilt} on hateful Memes~\cite{douwe2020hatefulmemes} dataset. Training with 100\% text + 100\% image; Testing with $\eta$\% text + 100\% image. Image only refers to the single modality setting (training and testing using image modality only). Reported values are averaged over 10 runs.}
%     \label{fig:hate}
% \end{figure}

\textbf{The ``missing'' test set.} We follow a conventional setting~\cite{tsai2018learning} to evaluate the model robustness against missing-modal data, in which the training data are modal-complete, while the testing data are modal-incomplete. We denote the full-modal train/test set as $100$\% Image + $100$\% Text and the missing-modal test set as $100$\% Image + $\eta$\% Text, where $\eta$\% is percentage of observed modality and $\eta$ indicates the severity of modality missing. The smaller the $\eta$, the severer the modality missing. When $\eta =0$, we evaluate the model's unimodal performance.  

% \begin{figure*}[ht]
%     \centering
%     \includegraphics[width=\linewidth]{figures/fig_policy_new22.pdf} 
%     \caption{\emph{Left}: Visualization of the learned policy. \emph{Right}: Example sample from MM-IMDb and Hateful Memes. \textit{Late fusion yields the best robustness on MM-IMDb, while early fusion leads to the most robust model on Hateful Memes.} The reported results were obtained using the following settings: training with 100\% Image + 100\% Text, testing with 100\% Image + 10\% Text.}
%     \label{fig:policy}
% \end{figure*}

\textbf{Robustness to missing modality.} We compare Transformer robustness on three multimodal datasets. Results are shown in Figure \ref{fig:imdb}. As shown, our method improves Transformer robustness on datasets with dominant modalities. Specifically, on \textit{MM-IMDb} and \textit{UPMC Food-101}, the model performance degrades as $\eta$ decreases. When text modality is severely missing, \textit{i.e.,} only 10\% are available, the multimodal performance is even worse than the unimodal one. For example, on MM-IMDb, the baseline model achieves an F1 Macro of $23.1$, which is $34.0$\% lower than the unimodal one of $35.0$; on UPMC Food-101, the baseline model achieves Accuracy of $58.4$ which is $18.3$\% lower than the unimodal Accuracy of $73.3$. However, in our method, the Transformer model still maintains good performance when modality is severely missing (only $10$\% text modality is observed), \textit{i.e.,} on MM-IMDb, our model yields F1 macro of $37.3$ which is $6.6$\% larger then unimodal F1 Macro of $35.0$; on UPMC Food-101, our method achieves an accuracy of $73.3$, which is $2.5$\% greater than the unimodal accuracy of $71.5$. 

Our method improves Transformer robustness on datasets with equally important modalities. Different from MM-IMDb and UPMC Food-101, \textit{the Hateful Memes} dataset does not have dominant modalities. We observe that the multimodal performance is always superior to the unimodal one. In this dataset, our method outperforms the baseline model when tested with modal-complete and modal-incomplete data. As shown in Figure \ref{fig:imdb}, when only $10\%$ of text is observed, our model yields an AUROC of $59.6$, which is $2.8$\% higher than the baseline ($58.0$); when full modalities are observed, our model outperforms the baseline by $2.3$\%.

\subsection{Analysis of Optimal Fusion Strategy}
\label{sec5:layer}
\begin{figure*}[ht]
    \centering
    \includegraphics[width=0.95\linewidth]{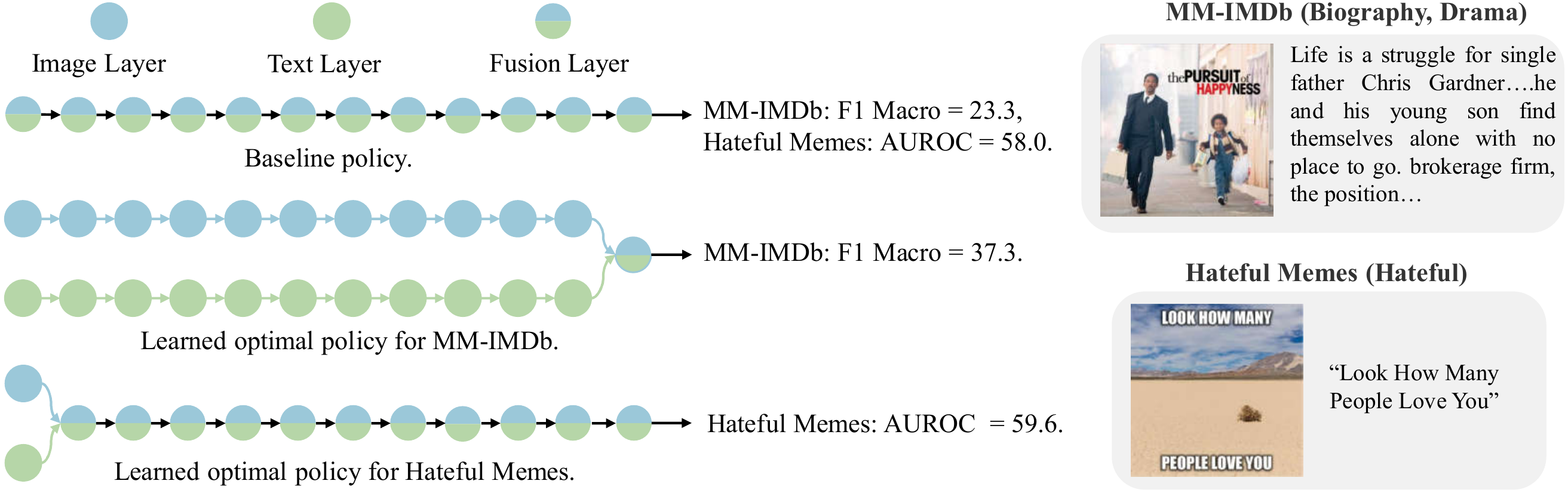} 
    \caption{\emph{Left}: Visualization of the learned policy. \emph{Right}: Example sample from MM-IMDb and Hateful Memes. \textit{Late fusion yields the best robustness on MM-IMDb, while early fusion leads to the most robust model on Hateful Memes.} The reported results were obtained using the following settings: training with 100\% Image + 100\% Text, testing with 100\% Image + 10\% Text.}
    \label{fig:policy}
\end{figure*}

% \textbf{The Optimal Policies}. The experiment results show that the optimal layers for multimodal fusion are different. 
We visualize the optimal policy of three datasets in Figure \ref{fig:policy}. We observe that late fusion is preferred on MM-IMDb, while early fusion is preferred on Hateful Memes. The learned policy is consistent with the characteristics of each dataset. Recall that in Sec.~\ref{subsec3:layer}, the depth of the fusion layer influences Transformer capacity to model cross-modality relations. The deeper the fusion layers, the lower the capacity. On MM-IMDb, the dominant modality text (plot descriptions) provides more details of the movie genres than the image modality (poster). It is reasonable that the model adopts a late fusion strategy since the prediction task can easily be addressed utilizing the dominant modality, and modeling cross-modal relations only brings marginal gains. In contrast, Hateful Memes dataset is constructed to fail models that rely on a single modality by adding challenging samples (“benign confounders”) to the dataset. Therefore, to handle this dataset, the model should have enough capacity to model cross-modal relations. For a dataset that relies on both modalities to make an accurate prediction, it is reasonable for our method to learn an early fusion strategy.

% The model need more capacity to learn the cross-modality relationships. We can gain more intuition by exploring the samples of MM-IMDb and Hateful Memes, as shown in Figure~\ref{fig:policy}. When it comes to identifying movie genres, the plot descriptions are often more accurate than the movie posters. 

\subsection{Ablation Study}
\label{sec5:ablation}
% \begin{figure}[ht]
%     \centering
%     \includegraphics[width=\linewidth]{figures/fig_4.pdf} 
%     \caption{Robustness of ViLT~\cite{kim2021vilt} on MM-IMDb~\cite{arevalo2017gated} dataset. Training with 50\% text + 100\% image; Testing with $\eta$\% text + 100\% image. Image only refers to the single modality setting (training and testing using image modality only).}
%     \label{fig:imdb2}
% \end{figure}
% We present ablation study to test the effectiveness of our method.

% \textbf{Comparison with other methods.} We conduct experiments using other simple method to improve the robustness of multimodal Transformer, \textit{i.e.,} random modality drop and naive imputation. We report the best performance in Table \ref{tab:ab1}. Compared with the two simple method, our multi-task learning-based method still gives the best result.
\noindent \textbf{Comparison with a new baseline.} Training with missing modalities is a simple way to improve model robustness. We implement this method as a new baseline. Results are shown in Table~\ref{tab:newbaseline}. Our experiments show that this simple method does not work. As shown, the performance of the new baseline is even worse than the unimodal baseline (Image only) on Food-101 and Hateful Memes.
\begin{table}[ht]
\vspace{-0.5\baselineskip}
\centering
\caption{Results of the new baseline: training and testing with 100\% image + 30\% text. }\label{tab:newbaseline}
\vspace{-0.5\baselineskip}
\resizebox{0.9\linewidth}{!}{\begin{tabular}{cccc}
\toprule
Method      & MM-IMDb & Food-101 & Hateful Memes \\ \midrule
Image only   &  31.2       &   65.9       &      60.2        \\ 
New baseline   &  40.4       &   44.3       &      59.7        \\ \midrule 
Ours  & \textbf{46.6}    & \textbf{77.5}     & \textbf{61.2}         \\ \bottomrule
\end{tabular}}
\end{table}

\noindent\textbf{Analysis on the multi-task learning and optimal fusion layers.} We conduct experiments to validate the effectiveness of each component under two different evaluation settings, \textit{i.e.,} $30$\% or $10$\% text are available. Results are shown in Table~\ref{tab:ab1}. Both components improve Transformer robustness. Furthermore, we find that multi-task learning contributes more than the fusion strategy. In detail, when only $10$\% of text is available at test, multi-task learning outperforms the optimal fusion policy by $30$\%.

\noindent\textbf{Analysis on the attention mask.} In our method, we apply masks on the attention matrix to enforce the classification tokens to leverage information only from the corresponding modalities. We study the effect of attention masks. The results are shown in Table \ref{tab:ab2}. We observe that it is important to make sure that each classification token is not peeking the information from other modalities.

% Additionally, it is interesting to see that, for the Transformer model, the domain modalities have large influence on the supportive modality. 

% \begin{table}[t]
% \centering
% \resizebox{\linewidth}{!}{
% \begin{tabular}{ccccccc}
% \hline
% \multicolumn{2}{c}{Method}  & \multicolumn{2}{c}{Training} & \multicolumn{2}{c}{Testing}       & \multirow{2}{*}{F1-Macro} \\ \cline{1-6}
% Multi-task & Opt. Policy & Image         & Text         & Image & Text                      &                           \\ \hline
%     \cmark       &  \cmark              & 100\%         & 100\%        & 100\% & \multicolumn{1}{c|}{30\%} & 41.8                      \\
%     \cmark       &  \cmark              & 100\%         & 100\%        & 100\% & \multicolumn{1}{c|}{10\%} & 37.3                      \\ \hline
%           &    \cmark            & 100\%         & 100\%        & 100\% & \multicolumn{1}{c|}{30\%} &                           \\
%           &    \cmark            & 100\%         & 100\%        & 100\% & \multicolumn{1}{c|}{10\%} &                           \\ \hline
%     \cmark       &                & 100\%         & 100\%        & 100\% & \multicolumn{1}{c|}{30\%} & 31.2                      \\
%     \cmark       &                & 100\%         & 100\%        & 100\% & \multicolumn{1}{c|}{10\%} & 22.6                      \\ \hline
% \end{tabular}}
% \caption{ Comparison with other straightforward methods for creating a robust multimodal Transformer on MM-IMDb.}\label{tab:ab1}
% \end{table}

\begin{table}[t]
\caption{ Ablation study on multi-task learning and optimal fusion on MM-IMDb.}\label{tab:ab1}
\centering
\resizebox{\linewidth}{!}{
\begin{tabular}{ccccccc}
\toprule
\multicolumn{2}{c}{Method} & \multicolumn{2}{c}{Training} & \multicolumn{2}{c}{Testing}       & \multirow{2}{*}{F1 Macro} \\ \cmidrule{1-6}
Multi-task   & Opt. Policy  & Image         & Text         & Image & Text                      &                           \\ \midrule
    \cmark         &             & 100\%         & 100\%        & 100\% & \multicolumn{1}{c|}{30\%} &   31.2                    \\
             &    \cmark         & 100\%         & 100\%        & 100\% & \multicolumn{1}{c|}{30\%} & 28.6                      \\
    \cmark         & \cmark            & 100\%         & 100\%        & 100\% & \multicolumn{1}{c|}{30\%} &  \textbf{41.8}                         \\ \midrule
    \cmark         &             & 100\%         & 100\%        & 100\% & \multicolumn{1}{c|}{10\%} & 22.6                          \\
             & \cmark            & 100\%         & 100\%        & 100\% & \multicolumn{1}{c|}{10\%} &    17.3                   \\
    \cmark         & \cmark            & 100\%         & 100\%        & 100\% & \multicolumn{1}{c|}{10\%} & \textbf{37.3}                      \\ \midrule
\end{tabular}}

\end{table}

\begin{table}[ht]
\caption{ Ablation study on the effect of attention mask for multi-task learning on MM-IMDb.}\label{tab:ab2}
\centering
\resizebox{\linewidth}{!}{
\begin{tabular}{cccccc}
\toprule
\multirow{2}{*}{Method}             & \multicolumn{2}{c}{Training} & \multicolumn{2}{c}{Testing}       & \multirow{2}{*}{F1 Macro} \\ \cmidrule{2-5}
                                     & Image         & Text         & Image & Text                      &                              \\ \midrule
\multicolumn{1}{c|}{without masking} & 100\%         & 100\%        & 100\% & \multicolumn{1}{c|}{10\%} & 23.0                        \\
\multicolumn{1}{c|}{with masking}    & 100\%         & 100\%        & 100\% & \multicolumn{1}{c|}{10\%} & 37.3                         \\ \bottomrule
\end{tabular}}
% \vspace{-4mm}
\end{table}

\section{Conclusion}
% We present a systematic study of the robustness of multimodal Transformer to missing modality. 
We empirically find that Transformer models are sensitive to missing-modal data. And surprisingly, the optimal fusion strategy is dataset-dependent; there does not exist a universal strategy that works in the presence of modal-incomplete data. Based on the findings, we build a robust Transformer via multi-task optimization. We develop an algorithm that automatically searches the optimal fusion strategies on different datasets. The searching for optimal fusion layers and network training are formulated into a bi-level optimization problem. Experiments across multiple benchmark datasets verify the superior robustness of our method. The limitation of our method is that multi-task learning can only ensure the multimodal performance is not worse than the unimodal one, which may not meet the requirements of safety-critical systems, such as autonomous driving. We plan to explore the effectiveness of generative-based methods, \textit{e.g.}, reconstructing the missing tokens to improve Transformer robustness against missing modality.

% \clearpage

\section*{Acknowledgements} This work is partially supported by NSF (CMMI-2039857) Research Grant and Snap Gift Research Grant.
% {\noindent\textbf{Acknowledgements.} This work is partially supported by NSF (CMMI-2039857) Research Grant and Snap Gift Research Grant.}

%%%%%%%%% REFERENCES
{\small
\bibliographystyle{ieee_fullname}
\bibliography{references}
}
% \clearpage
% \input{sec7_supp}
\end{document}